\documentclass[11pt]{article}
\usepackage{coling2020}
\usepackage{times}
\usepackage{url}
\usepackage{latexsym}

\usepackage{times}
\usepackage{latexsym}
\usepackage{microtype}
\usepackage{xspace}
\usepackage{url}
\usepackage{booktabs}
\usepackage{adjustbox}
\usepackage{multirow}

\usepackage{makecell}
\usepackage{graphicx}
\usepackage{subcaption}
\usepackage{color, colortbl}

\usepackage[shortlabels]{enumitem}

\usepackage{caption}
\usepackage{arydshln}

\usepackage{amssymb}
\usepackage{amsmath}
\usepackage{amsfonts}
\usepackage[title]{appendix}
\usepackage{amsmath}

\usepackage{pifont}

\newcommand{\word}[1]{\textit{#1}}
\newcommand{\inflect}[1]{\textit{#1}}
\newcommand{\lexeme}[1]{\textsc{#1}}

\newcommand{\defn}[1]{\textbf{#1}}
\newcommand{\model}[1]{\textsc{#1}}

\newcommand{\lang}[1]{\textsc{#1}}

\definecolor{amber}{rgb}{0.91, 0.41, 0.17}
\newcommand{\R}[1]{{\color{black} #1}}
\newcommand{\B}[1]{{\color{black} #1}}
\newcommand{\W}[1]{{\textbf{#1}}}

\newcommand{\vpi}{{\boldsymbol \pi}}

\newcommand{\reals}{\mathbb{R}}

\newcommand{\calR}{R}
\newcommand{\embed}{\mathbf{e}}

\newcommand{\T}[1]{\textsc{\MakeLowercase{#1}}}

\usepackage{cleveref}
\crefname{section}{\S}{\S\S}
\Crefname{section}{\S}{\S\S}
\crefname{table}{Table}{}
\crefname{figure}{Figure}{}
\crefname{algorithm}{Algorithm}{}
\crefname{equation}{eq.}{}
\crefname{appendix}{Appendix}{}
\crefformat{section}{\S#2#1#3}

\DeclareMathOperator*{\argmaxA}{arg\,max}

\definecolor{ghostwhite}{rgb}{0.95, 0.95, 0.96}

\colingfinalcopy 

\title{Morphologically Aware Word-Level Translation}

\newcommand{\ucambridge}{\normalfont 1}
\newcommand{\ethz}{\normalfont 2}
\newcommand{\deepmind}{\normalfont 3}

\newcommand\tab[1][1cm]{\hspace*{#1}}

\author{Paula Czarnowska$^{\ucambridge}$ \tab[0.5cm] Sebastian Ruder$^{\deepmind}$ \tab[0.5cm]
\textbf{Ryan Cotterell$^{\ucambridge, \ethz}$ \tab[0.5cm] Ann Copestake$^{\ucambridge}$} \\
  $^{\ucambridge}$University of Cambridge \tab[0.5cm] $^{\deepmind}$DeepMind \tab[0.5cm]
  $^{\ethz}$ETH Z\"{u}rich \\
  \texttt{pjc211@cam.ac.uk}~\;~  \texttt{sebastian@ruder.io} \\
  \texttt{ryan.cotterell@inf.ethz.ch}~\;~ \texttt{aac10@cam.ac.uk}
}

\date{}

\begin{document}
\maketitle
 
 \begin{abstract}
 We propose a novel morphologically aware probability model for bilingual  lexicon induction, which jointly models lexeme translation and inflectional morphology in a structured way.
 Our model exploits the basic linguistic intuition that the lexeme is the key lexical unit of meaning, while inflectional morphology provides additional syntactic information. 
 This approach leads to substantial performance improvements---19\% average improvement in accuracy across 6 language pairs over the state of the art in the supervised setting and 16\% in the weakly supervised setting.
 As another contribution, we highlight issues associated with modern BLI that stem from ignoring inflectional morphology, and propose three suggestions for improving the task.
\end{abstract}

\section{Introduction} 
The ability to generalize to rare and unseen morphological variants of a known word lies at the heart of translation. 
For instance, a capable human Spanish--English translator would find translating the exceptionally rare form \word{tosed} (2$^{\text{nd}}$ person plural imperative form of `to cough') as straightforward as translating the infinitive \word{toser}---despite the fact that \word{tosed} is so infrequent that many native Spanish speakers may have never encountered the form themselves. 
Given how basic this generalization ability is for humans, one should expect a good bilingual lexicon inducer to exhibit a similar capacity to generalize. 
In other words, the model should translate infrequent, regular forms of a lexeme as accurately as it translates a lexeme's most common forms.\looseness=-1

Nevertheless, current approaches to bilingual lexicon induction (BLI) fall short of this desideratum. 
In a recent study, \newcite{czarnowska2019dont} reveal that the performance of state-of-the-art bilingual lexicon inducers degrades severely when translating less frequent inflected forms---even for the most common lexemes. 
The problem is severe: In the case of inducing a French--Spanish bilingual lexicon, the model of \newcite{ruder2018discriminative} correctly translates infinitives 50.6\% of the time, but correctly translates the 2$^{\text{nd}}$ person plural imperative forms only 1.5\% of the time. 
Motivated by this disparity, this work introduces a novel morphologically aware probability model for BLI that jointly models lexeme translation and inflectional morphology in a structured way. 
Our model exploits the basic linguistic intuition that the lexeme is the core lexical unit of meaning, while inflectional morphology provides additional syntactic information on top of it \cite{haspelmath2013understanding}.
It follows that we should ignore this syntactic information when translating at the word level and handle morphological inflection with a different component of the model.

The empirical portion of our paper describes experiments on French, Italian and Spanish. 
We find our joint model substantially improves over several strong baselines on the BLI task. 
When evaluating on held-out lexemes,\footnote{This is the case when we have never seen any inflected form of a given lexeme in the training data.}
we observe an average performance improvement of 19\% and 16\% over the previous state of the art \cite{artetxe2018robust} in the supervised and weakly supervised settings, respectively. 
In addition, we propose a simple heuristic to further boost performance: Inspired by the dual-route hypothesis \cite{pinker1994regular,pinker1998words} and the network model of morphological processing \cite{bybee1985morphology,Bybee1995Regular}, we translate high-frequency forms, which are most likely to exhibit irregularity, directly (without going through the lexeme) and reserve our morphologically aware model for low-frequency forms. This heuristic gives us a further 2\% improvement.

\section{Background}\label{sec:background}

\subsection{Morphological Inflection}
Morphological inflection is the systematic alteration of the word form that adds specific morpho-syntactic information to the lexeme, e.g. tense, case and number. 
English is weakly inflected with only a few forms per lexeme, and, in that respect, it differs from many other languages.
A higher degree of inflection is indeed the norm among the world's languages \cite{dryer2013world}. 
In this work, we distinguish the terms lexemes, inflected forms and lemmata. 
A \defn{lexeme} is an abstract concept that represents the core meaning shared by a set of inflected forms. 
An \defn{inflected form} is an individual morphological variant that belongs to a given lexeme. 
As an example, the lexeme \lexeme{run} has the inflected forms \inflect{run}, \inflect{runs}, \inflect{ran}, and \inflect{running}. 
The \defn{lemma}, also called the citation form, is an inflected form that lexicographers have chosen to be representative of the lexeme. 
For example, the lexeme \lexeme{run}'s lemma is \inflect{run}. 
In many languages, the infinitive is the verbal lemma and the nominative singular is the nominal lemma.
We consider a \defn{lexicon} of a language to be a set of inflected forms.\looseness=-1\footnote{This full-listing view of the lexicon is taken, among others, by \newcite{jackendoff1975morphological}, \newcite{mcclelland1987parallel} and \newcite{Bybee1995Regular},
and stands in opposition to alternative views of the lexicon being comprised of only the unpredictable items (e.g. \newcite{prasada1993generalisation} \newcite{pinker1994regular}).}

\subsection{Bilingual Lexicon Induction}
In the NLP literature, the BLI task is to translate a given list of source-side word forms into the most appropriate corresponding target-side word forms. 
It dates back to 1990s. The first data-driven experiments on parallel corpora  made use of word-alignment techniques \cite{brown-etal-1990-statistical,kupiec1993,smadja1994translating}. 
 Such approaches were later extended to operate on non-parallel or even unrelated texts by leveraging the correlation between word co-occurrence patterns in different languages \cite{rapp1995identifying,fung1998translating,Fung1998,koehn2002learning}. Apart from the distributional signal, the early approaches make use of other monolingual clues, e.g. word spelling, cognates or word frequency.\footnote{For more references we refer the reader to the survey of \newcite{ruder2019survey}.} 
More recent approaches leverage the distributional signal in word embeddings without any explicit linguistic clues.
Many current models \cite{mikolov2013exploiting,ruder2019survey} learn a linear transformation between two monolingual word embedding spaces, often guided by an initial set of seed translations.
{ This seed dictionary frequently spans several thousand word pairs \cite{mikolov2013exploiting,xing2015normalized,lazaridou2015hubness,artetxe2016learning} but one can also provide weaker supervision, through listing only identical strings or shared numerals \cite{artetxe2017learning,Søgaard2018}}. 
For unsupervised BLI, the initial translations may also be induced automatically through exploiting the structure of the monolingual embedding spaces \cite{zhang2017adversarial,conneau2018word,artetxe2018robust}.
We focus on supervised and weakly supervised BLI which outperform unsupervised approaches \cite{Glavas2019}.
The BLI models are typically evaluated using the precision@$k$ metric, which tells us how many times the correct translation of a source form is among the $k$-best candidates returned by the model.
In this work we exclusively consider the precision@1 metric, which is the least forgiving.

\subsection{Morphological Inflection: A Challenge for BLI}\label{sec:rendez-vous}
Most datasets for BLI operate at the level of inflected forms and impose no restriction on the morpho-syntactic category of translated words. 
From a lexicographer's standpoint, this choice is unusual. 
Dictionaries generally only list lemmata; inflected forms are rarely listed. Thus, BLI is \emph{not} the task of inducing a dictionary in the strict sense of the word. 
The authors have found this a common misconception in the literature and among NLP researchers.

Despite the assumption that inflectional morphology is present in the lexicon, most BLI datasets list only a handful of inflected forms per lexeme. 
This is due to frequency {restrictions imposed when the datasets are created} and the consequences can be quite severe; \newcite{czarnowska2019dont} reveal that the \model{muse} test dictionaries \cite{conneau2018word} for Romance language pairs cover, on average, only 3\% of paradigms for verbs they contain. 
In contrast, we assert that BLI models should be trained and evaluated on datasets that contain a more representative range of morphological inflections. We use the term \defn{morphologically enriched dictionary} for such bilingual lexicons (see \cref{sec:datasets}).

To our knowledge, we are the first to explicitly model inflectional morphology in BLI. Closest to our endeavor,
\newcite{Yang2019} address morphology in BLI by incorporating grammatical information learned by a pre-trained denoising language model, while \newcite{riley2018orthographic} enhance the projection-based approach of \newcite{artetxe2017learning} with orthographic features to improve performance on BLI for related languages.

\section{A Joint Model for Morphologically Aware Word-level Translation}\label{sec:joint-model}
The primary contribution of this work is a morphologically aware probabilistic model for word-level translation. 
Our model exploits a simple intuition: Because the core unit of meaning is the lexeme, one should translate \emph{through} the lexeme and then inflect the word according to the target language's morphology.

\paragraph{Notation.} 
In the task of BLI, we consider a source language $s$ and a target language $t$. The goal is to translate inflected forms in a source language $\iota_s \in L_s$ to inflected forms in a target language $\iota_t \in L_t$, where $L_s$ and $L_t$ are the source-side and the target-side lexicons, respectively. 
We denote source-side lemmata as $\lambda_s \in L_s$ and target-side lemmata as $\lambda_t \in L_t$. 
We use $\tau_s \in T_s$ for the morpho-syntactic tag of the source form $\iota_s$ and $\tau_t \in T_t$ for the tag of the target form $\iota_t$, where $T_s$ and $T_t$ are the sets of possible tags in the source and the target language. 
Finally, let $\embed : L \rightarrow \reals^N$ be a function which takes as input a word and returns its pre-trained monolingual word embedding. 

\paragraph{Model.} 
We construct a joint probability model for morphologically aware word-level translation. 
Using the notation defined in the previous paragraph, we formally define the model as
\begin{align}\label{eq:joint-model}
&p(\vpi_t, \lambda_t,\tau_s, \lambda_s,  \mid \iota_s) = \underbrace{\left[\prod_{\langle\iota_t, \tau_t \rangle \in \vpi_t} \!\!\!\!p(\iota_t, \tau_t \mid \lambda_t, \tau_s)\right]}_\textit{synthesizer} \underbrace{p(\lambda_t \mid \lambda_s)}_{\textit{translator}} \underbrace{p(\tau_s, \lambda_s  \mid \iota_s)}_\textit{analyzer} 
\end{align}
where $\vpi_t$ is a \emph{set} of valid target-language translations. The joint distribution
is factorized into three parts: a synthesizer, a translator and an analyzer. 
In the next three sections, we define each of these
distributions. 
Note that the model, as defined in \cref{eq:joint-model}, can provide a \emph{set} of valid translations $\vpi_t$. 
For languages with very similar morphological systems, this set will often have one element (with the same morpho-syntactic description),  while for more distinct languages it will contain a number of inflected forms. 

\subsection {The Synthesizer: $\prod\limits_{\langle\iota_t, \tau_t \rangle \in \vpi_t} \!\!\!\!p(\iota_t, \tau_t \mid \lambda_t, \tau_s)$}

The synthesizer produces a {set} of valid target-side inflected forms and their tags $\vpi_t$ given a target-side lemma $\lambda_t$ and a source-side tag $\tau_s$.  We formally define this distribution as follows:
\begin{align}
\prod\limits_{\langle\iota_t, \tau_t \rangle \in \vpi_t} p(\iota_t, \tau_t \mid \lambda_t, \tau_s) = \prod\limits_{\langle\iota_t, \tau_t \rangle \in \vpi_t} \!\!\!\! \underbrace{p(\iota_t \mid \tau_t, \lambda_t)}_\textit{inflector}\ \underbrace{p(\tau_t \mid \tau_s)}_\textit{tag translator}
\end{align}
The joint distribution over forms and tags is factored into two parts. 
The first part, the inflector, produces an inflected form $\iota_t$ given a lemma $\lambda_t$ and a morphological tag $\tau_t$. 
This problem has been well studied in the NLP literature \cite{cotterell-etal-2016-sigmorphon,Cotterell2017CoNLLSIGMORPHON}.
The second part, tag translator, determines the possible target-side morphological tags that are compatible with the features present in the source tag. 

In principle, our model is compatible with any probabilistic inflector. 
In this paper, we employ the model of \newcite{wu2019exact}, which obtained the single-model state of the art at the time of experimentation \cite{mccarthy2019sigmorphon}. 
The model has a latent character-level monotonic alignment between the source and target inflections that is jointly learned with the transducer and is, in effect, a neuralized version of a hidden Markov model for translation \cite{vogel-etal-1996-hmm}. 

In this work we focus on closely related languages and make a simplifying assumption that there exists a single most-plausible translation for each inflected form.\footnote{Indeed, the dictionaries of \newcite{czarnowska2019dont} on which we experiment also make this assumption.} We formalize the tag translator using an indicator function:
\[
 p(\tau_t \mid \tau_s) = 
  \begin{cases} 
   1 & \textbf{if } \tau_t = \tau_s \\
   0 & \textbf{if } \tau_t \neq \tau_s
  \end{cases}
\]
For experiments with more distant language pairs one can define $p(\tau_t \mid \tau_s)$ to be a multi-label classifier.

\subsection{The Translator: $p(\lambda_t \mid \lambda_t)$}
As our translator, we construct a log-bilinear model that yields a distribution over all elements in the target lexicon. 
We assume the existence of both source- and target-side
embeddings. 
For notational simplicity, we use $\embed$ to define the embedding function for both the source and the target language, although in practice these look-up functions are distinct.
The model has a single matrix of parameters: $\mathbf{\Omega} \in \reals^{N_t \times N_s}$ where $N_s$ is the source embedding dimensionality and $N_t$ the target embedding dimensionality. 
Our translator is defined as the following conditional model
\begin{equation}
    p(\lambda_t \mid \lambda_s) = \frac{1}{Z(\lambda_s)} \exp \big(\mathbf{e}(\lambda_t)^{\top}\ \mathbf{\Omega}\ \mathbf{e}(\lambda_s) \big)
    \label{eq:translator}
\end{equation}
where the normalizer is defined as
\begin{equation}
    Z(\lambda_s) = \sum_{\lambda'_{t} \in L_t} \exp \big(\mathbf{e}(\lambda'_{t})^{\top}\ \mathbf{\Omega}\ \mathbf{e}(\lambda_s) \big)
    \label{eq:translator2}
\end{equation}
{Note that this log-bilinear model differs from most
 embedding-based bilingual lexicon inducers which predict embeddings, rather than words.}
For example, \newcite{ruder2018discriminative}'s approach contains a multivariate Gaussian over the target-language's embedding space.\footnote{Note that the translator's distribution, as defined in \cref{eq:translator} and \cref{eq:translator2}, is over all inflected forms in the target lexicon. 
An alternative would be to define a distribution over lemmata only. 
However, this would require filtering out all non-lemma forms from the target embedding matrix, which is not trivial. 
In our preliminary experiments, we observed that this can lead to a further performance increase.}

\paragraph{Orthogonal Regularization.}
During training we employ a special regularization term on the parameter matrix $\mathbf{\Omega}$. Specifically, we use
\begin{equation}
\calR(\mathbf{\Omega}) = \alpha \left|\left|\mathbf{\Omega}^\top \mathbf{\Omega} - \mathbf{I} \right|\right|_{\mathrm{F}}
\end{equation}
\noindent with a tunable ``strength'' hyperparameter {$\alpha \in \reals_{\geq 0}$}.
This term encourages the translation matrix to be orthogonal, which has led to consistent gains in past work \cite{xing2015normalized,artetxe2016learning,ruder2018discriminative}.

\subsection{The Analyzer: $p(\lambda_s, \tau_s \mid \iota_s)$}
{For our probabilistic analyzer we use the same hard attention model as in the inflector. The model predicts both the lemma and the morphological tag; the output is a morphological tag followed by a special end-of-tag character and a sequence of lemma characters.}

\section{A Frequency-Based Heuristic} \label{sec:hybrid}
As defined in \cref{sec:joint-model}, the model handles regular and irregular word forms in the same manner. 
This can be problematic, as the analysis and synthesis modules have no means of handling irregular morphology, beyond the irregular forms they have been exposed to during training. 
Guided by this insight, we propose an alternative version of our model, which employs a special treatment for forms likely to have irregular morphology---the most frequent forms \cite{bybee1995diachronic,baayen1996word,Wu2019Morphological}.
We term this extension the \defn{hybrid model}. 
It employs a frequency-based heuristic and translates the source form through its lemma \emph{only} if the lemma is more frequent.\footnote{We rely on the order of \model{fastText} embeddings for the relative ranking of inflected forms.} 
Otherwise, it translates the inflected form directly, using only the translator component. 
For example, the Spanish form \word{pide}---an irregular inflected form of \word{pedir} (`to ask for')---would be translated directly, since \word{pedir} is less frequent.

In a broader context of morphological processing, different handling of a form depending on its frequency or regularity can be linked to the dual-route hypothesis \cite{pinker1994regular,pinker1998words},{ which posits that regular and irregular inflection are handled by different cognitive mechanisms}, or the works of \newcite{Baayen1992}, \newcite{Baayen1993} and \newcite{hay2001lexical}, which have loosely inspired this heuristic.\footnote{In particular, \newcite{hay2001lexical} advocates the importance of \emph{relative} (vs. absolute) frequency for lexical access, arguing that it is more beneficial to define a frequent complex form as one that is more frequent than its base, regardless of how often it is used.}

\section{Experimental Setup}
Our evaluation involves 3 Romance languages which exhibit a higher degree of inflection and are commonly experimented on within BLI---French, Spanish and Italian. 
Because these languages come from the same branch of the Indo-European family, the results serve as an empirical upper-bound on BLI.

\subsection{BLI Datasets} \label{sec:datasets}
\paragraph{\newcite{czarnowska2019dont}.}
As discussed in \cref{sec:rendez-vous}, given that inflectional morphology is present in the induced lexicon, BLI models should be trained and evaluated on datasets which list a range of compatible inflected form pairs for every source-target lexeme pair. 
At this time, the dictionaries of \newcite{czarnowska2019dont} are the only publicly available resource that meets this criterion, and, for this reason, they are the most important evaluation benchmark used in this work.
The dictionaries were generated based on Open Multilingual WordNet \cite{bond2012survey}, Extended Open Multilingual WordNet \cite{bond2013linking} and UniMorph\footnote{\url{https://unimorph.github.io/}} \cite{kirov2016very,mccarthy-etal-2020-unimorph}, a resource comprised of inflectional word paradigms for 107 languages.
The dictionaries only list parts of speech that undergo inflection in either the source or the target language; these are nouns, adjectives and verbs in the Romance languages.

\paragraph{\newcite{conneau2018word}.}
\model{muse} \cite{conneau2018word} was generated using an ``internal translation tool'' and is one of the few other resources which covers pairs of Romance languages. 
However, it is skewed towards most frequent forms: The vast majority of forms in \model{muse} are ranked in the top 10k of the vocabularies in their respective languages, causing it to omit many morphological variants of words. The dataset also suffers from other issues, such as a high level of noise coming from proper nouns \cite{kementchedjhieva2019lost}. 
Thus, we do not view this resource as a reasonable benchmark for BLI.

\subsection{Baselines}

\paragraph{\newcite{artetxe2016learning}.}
They learn an orthogonal linear transformation matrix between the source language space and the target language space, after length-normalizing and mean-centering the monolingual embedding matrices. 
Their method is fully supervised and works best with large amounts of training data (several thousand translation pairs).

\paragraph{\newcite{ruder2018discriminative}.}
They introduce a weakly supervised, self-learning model, which can induce a dictionary given only a handful of initial, seed translations. 
This is achieved by iteratively alternating between two steps: an inflected form alignment step and a mapping step. 
The first is comprised of finding a matching in a bipartite weighted graph in which source forms constitute one set and target forms the other set. 
During the mapping step the resulting alignment is used to learn a better projection from the source language space into the target space---this is done by solving the orthogonal Procrustes problem.\footnote{{Given two matrices $\mathbf{A}$ and $\mathbf{B}$, the orthogonal Procrustes problem is to find an orthogonal matrix $\mathbf{Q}$ which most closely maps from $\mathbf{A}$ to $\mathbf{B}$: $\mathbf{Q} = \argmaxA_{\mathbf{\Omega}} \left\| \mathbf{\Omega} \mathbf{A} - \mathbf{B} \right\|_\mathrm{F} $ subject to $\mathbf{\Omega}^{\top} \mathbf{\Omega} = \mathbf{I}$}}

\paragraph{\newcite{artetxe2018robust}.} 
They propose a fully unsupervised approach to BLI.
The starting point for their model is an automatic initialization of a seed dictionary which exploits the structural similarity of the monolingual embeddings.
Like \newcite{ruder2018discriminative}, their model also utilizes self-learning, but learns mappings in both directions and employs a range of additional training techniques. 
Despite being unsupervised, this model constitutes
the state of the art on many BLI datasets (e.g. \newcite{dinu2015improving}'s and \newcite{Artetxe2018GeneralizingAI}'s), outperforming even the supervised approaches. 
Thus, we include it in our evaluation and compare it to supervised and semi-supervised approaches.\looseness=-1 

\subsection{Skyline}
We also consider a version of our model which uses an oracle analyzer---the source lemma $\lambda_s$ and  tag $\tau_s$ are known \textit{a priori}.
The skyline provides an upper-bound of performance---to wit, what performance would be achievable if the model had had access to more information about the translated source form.

\subsection{Experimental Details} \label{exp:setting}
We implemented all models in PyTorch \cite{NEURIPS2019_9015}, adapting the code of  \newcite{wu2019exact} for the transducers (analyzer and inflector). Throughout our experiments we used the Wikipedia \model{fastText} embeddings \cite{grave2018learning},{ which we length-normalized and mean-centered before training the models}. As is standard, we trained all translators on the top 200k most frequent word forms in the vocabularies of both languages. 
To evaluate on very rare forms present in the dictionaries of \newcite{czarnowska2019dont} which are out-of-vocabulary (OOV) for \model{fastText}, we created an OOV \model{fastText} embedding for every OOV form that appears in a union of WordNet and UniMorph and appended those representations to the original embedding matrices.\footnote{The \model{fastText} framework allows for creating a vector of an unseen form by summing the vectors of all character $n$-grams present in that form.} 
We evaluated all models using precision@1 as a metric, which is equivalent to accuracy. 
At evaluation, for all models we used cosine as a measure of similarity between two word embeddings.\footnote{CSLS \cite{conneau2018word} is an alternative retrieval method and can lead to better results.}

\paragraph{Estimating the Model.}
We estimate the parameters of our models to maximize the log-likelihood of the training data. In the supervised case, we are able to estimate the parameters of the different components of the model independently. For every language pair we trained a{ separate translator, on the initial seed dictionary, as well as a separate analyzer and inflector} on UniMorph entries---in the source language for the analyzers and the target language for the inflectors.
Importantly, to ensure that the transducers are never trained on forms they will see at evaluation, we excluded the entries present in the test or development split of the considered evaluation dataset.\footnote{Since the \model{muse} dataset contains word forms associated with parts of speech not present in UniMorph, i.e. adverbs, determiners and pronouns; in our experiments on those datasets we additionally enhanced the transducers' training data with forms associated with those POS. 
All these additional forms are base forms and the models are trained to analyze them as such and not to inflect them.}

For inflection, we make the assumption that there always exists one unique inflected form for a given lemma and tag. 
However, in the case of analysis, due to syncretism, there often exists a number of plausible interpretations. As a result, in the training data there might be a number of correct lemma--tag combinations for every input form. 
At training, we select only one of those possible analyses. 
We found that this approach works better than training the model on all possible analyses as targets. 
As a consequence, the analyzers might be biased towards specific morpho-syntactic interpretations of syncretic forms, which, down the line, may hurt the performance of our model. 
Indeed, we view this as a trade-off between having a more accurate but biased analyzer that limits the possible translations produced by the model, and a more noisy analyzer which can result in more varied translations.

\paragraph{Hyperparameters.}
The models were trained with Adam \cite{Kingma2014AdamAM}, using an initial learning rate of $0.001$ for the transducers, and $0.05$ for most translators (see \cref{sec:hyper-params} for a detailed breakdown). 
We halved the learning rate after every epoch for which the development loss increased and utilized early stopping (with the min. learning rate of $1 \times 10^{-8}$). 
For the transducers we also applied gradient clipping with a maximum gradient norm of 5. 
For \newcite{ruder2018discriminative}, we set the number of candidate target words considered for each source word during matching to 15 and constrained the matching to consider only 40k most frequent forms.

\vspace{2cm}
\begin{table*}
\centering
\renewcommand{\arraystretch}{1.3}
\hskip-0.2cm
\resizebox{0.8\textwidth}{!}{%
\begin{tabular}{  l | l l | l l | l l | l l | l l | l l   }
\toprule 
\multirow{2}{*}{Model} & \multicolumn{2}{ c |  }{\lang{fra--ita}} & \multicolumn{2}{ c |  }{\lang{fra--spa}} & \multicolumn{2}{ c |  }{\lang{ita--fra}} & \multicolumn{2}{ c |  }{\lang{ita--spa}} & \multicolumn{2}{ c |  }{\lang{spa--fra}} & \multicolumn{2}{ c   }{\lang{spa--ita}} \\ 
& \textsc{voc} & \textsc{all} & \textsc{voc} & \textsc{all} & \textsc{voc} & \textsc{all} & \textsc{voc} & \textsc{all} & \textsc{voc} & \textsc{all} &\textsc{voc} & \textsc{all} \\ 
\hline
  \multicolumn{13}{c}{\multirow{1}{*}{\textbf{Experiments on the morphologically enriched dictionaries \cite{czarnowska2019dont}}}} \\
\hline
 Train/test split sizes & \multicolumn{2}{c | }{94k/21k}  & \multicolumn{2}{c |}{85k/22k}  & \multicolumn{2}{c | }{96k/31k} & \multicolumn{2}{c |}{70k/31k}  &  \multicolumn{2}{c |}{86k/26k} &  \multicolumn{2}{c }{70k/24k} \\ 
 \hline
  \multicolumn{13}{c}{ Supervised setting: full train dictionary splits were used for training.} \\
\hline
 \newcite{artetxe2016learning} & 49.4 & 28.3 & 49.1 & 29.3 & 40.8 & 21.4 & 43.6 & 22.6 & 46.9 & 25.8 & 47.0   & 27.8 \\
 \newcite{ruder2018discriminative}   & 48.3 & 26.8 & 47.8 & 26.7 & 40.5 & 21.0   & 42.7 & 21.1 & 45.5 & 24.2 & 46.6 & 27.1 \\
 \R{Our base model} & 50.5 & 47.8 & 48.6 & 45.1 & 42.8 & 35.2 & 46.0 & 38.6 & 51.5 & 44.3 & 55.4 & 51.6 \\ 
  Our hybrid model   & \W{56.7} & \W{50.6}  & \W{54.6} & \W{47.9} & \W{47.2} & \W{36.9} & \W{50.3} & \W{40.2} & \W{55.2} & \W{45.9} & \W{58.3} & \W{52.9} \\

 \rowcolor{ghostwhite} \B{Oracle} & 58.1 & 52.2 & 56.3 & 49.3 & 48.7 & 37.3 & 51.8 & 41.0 & 56.7 & 46.8 & 61.2 & 54.3 \\
  
\hline
 \multicolumn{13}{c}{Weakly supervised setting: pairs of forms identical in both languages were used as a seed dictionary.} \\
\hline
 \newcite{artetxe2016learning} & 44.7 & 25.2 & 45.2 & 24.8 & 35.8 & 18.5 & 40.8 & 19.5 & 42.2 & 22.2 & 44.1 & 24.9 \\
 \newcite{ruder2018discriminative}   & 48.4 & 26.9 & 47.9 & 26.6 & 40.7 & 20.9 & 42.7 & 21.0   & 45.6 & 24.2 & 46.4 & 26.9 \\
 \R{Our base model} & 46.7 & 43.2 & 45.8 & 42.0 & 41.1 & 33.8 & 44.7 & 36.4 & 49.2 & 40.8 & 52.9 & 47.3 \\
  Our hybrid model  & \W{52.2} & \W{45.7}  & \W{51.2} & \W{44.3} & \W{44.7} & \W{35.2} & \W{48.9} & \W{37.9} & \W{52.5} & \W{42.3} & \W{55.8} & \W{48.7} \\
 \rowcolor{ghostwhite} \B{Oracle} & 53.6 & 47.0 & 53.0 & 45.6 & 46.9 & 36.0 & 50.3 & 38.6 & 54.3 & 43.6 & 58.3 & 50.1 \\
 \hline
 \multicolumn{13}{c}{Unsupervised setting: no seed dictionary was provided.} \\ \hline
   \newcite{artetxe2018robust} & 49.4 & 28.2 & 49.2 & 27.4 & 38.5 & 20.0 & 40.6 & 20.0 & 43.2 & 23.0 & 47.4 & 28.2
   \\
\hline
  \multicolumn{13}{c}{\multirow{1}{*}{\textbf{Experiments on the MUSE \cite{conneau2018word} (Train/test split sizes: 5000/1500 )}}} \\
\hline
  \multicolumn{13}{c}{Supervised setting: full train dictionary splits were used for training.} \\
\hline
 \newcite{artetxe2016learning} & \W{78.9} & \W{78.9} & \W{78.9} & \W{78.9} & 83.5 & 83.5 & \W{84.6} & \W{84.6} & 82.1 & 82.1 & \W{79.5} & \W{79.4} \\
 \newcite{ruder2018discriminative}   & 76.4 & 76.4 & 76.6 & 76.6 & \W{83.9} & \W{83.9} & 83.7 & 83.7 & \W{82.5} & \W{82.5} & 78.9 & 78.9 \\
Our base model   & 58.2 & 58.2 & 50.7 & 50.7 & 63.1 & 63.1 & 61.7 & 61.7 & 58.7 & 58.7 & 56.1 & 56.1 \\
 Our hybrid model & 75.1 & 75.1 & 71.6 & 71.6 & 79.7 & 79.7 & 81.4 & 81.4 & 75.1 & 75.1 & 75.4 & 75.3 \\
 \hline
 \multicolumn{13}{c}{Unsupervised setting: no seed dictionary was provided.} \\
\hline
\newcite{artetxe2018robust}  & 73.1 & 73.1 & 74 .0 & 74.0 & 77.7 & 77.7 & 81.0 & 81.0 & 77.7 & 77.7 & 76.6 & 76.6 \\
\bottomrule 
\end{tabular}
}
\caption{BLI results:
\textsc{voc} considers in-vocabulary forms and \textsc{all} considers the full dictionary, including OOVs.}
\label{tab:big-resuls}
\end{table*}

\paragraph{Decoding.}
We decode the model with greedy search, i.e. beam search with a beam size of 1. In all experiments we return the single most suitable form.

\paragraph{Data Requirements.}
As for other projection-based BLI approaches, the translator component needs to be trained on monolingual embeddings and an initial seed dictionary, which can be generated automatically. 
The \emph{only} additional resource required to train the full model is UniMorph, or a similar morphological database, which is used to train the transducers.
Although this extra requirement could be a limitation in some cases, such morphological lexicons are available in an increasingly large number of languages because they are a by-product of descriptive linguists' efforts to document the world's languages. 

\section{Results}\label{sec:results}

We consider the fully supervised and the weakly supervised setting. 
Weak supervision in the case of BLI refers to populating the seed dictionary with identically spelled strings \cite{Søgaard2018}. In \cref{tab:big-resuls}, we present our experimental results on the whole evaluation dictionary, including the words which are out-of-vocabulary for \model{fastText} (\textsc{all}), as well as on the in-vocabulary forms only (\textsc{voc}). 
When evaluated on the morphologically enriched resource, our proposed approach leads to substantial performance gains over the baseline models, on every language pair. 
In the supervised setting, for the in-vocabulary words we note an average 3\% improvement in the case of our base model and a 8\% improvement in the case of our hybrid model over the best performing baseline \cite{artetxe2016learning}. In the full-vocabulary experiments, the improvements reach 18\% for our base model and 20\% for the hybrid. 
In the weakly supervised setting, the performance gains remain similarly high---for our base model they reach 2\% and 16\%, respectively, compared to the best baseline \cite{artetxe2018robust}, while for the hybrid model we observe a 6\% and 18\% improvement, respectively. 
Our experiments with an oracle analyzer demonstrate that even larger gains are possible: the oracle consistently outperforms every other model across all evaluation conditions. 

\begin{figure*}
    \centering
    \includegraphics[height=7cm]{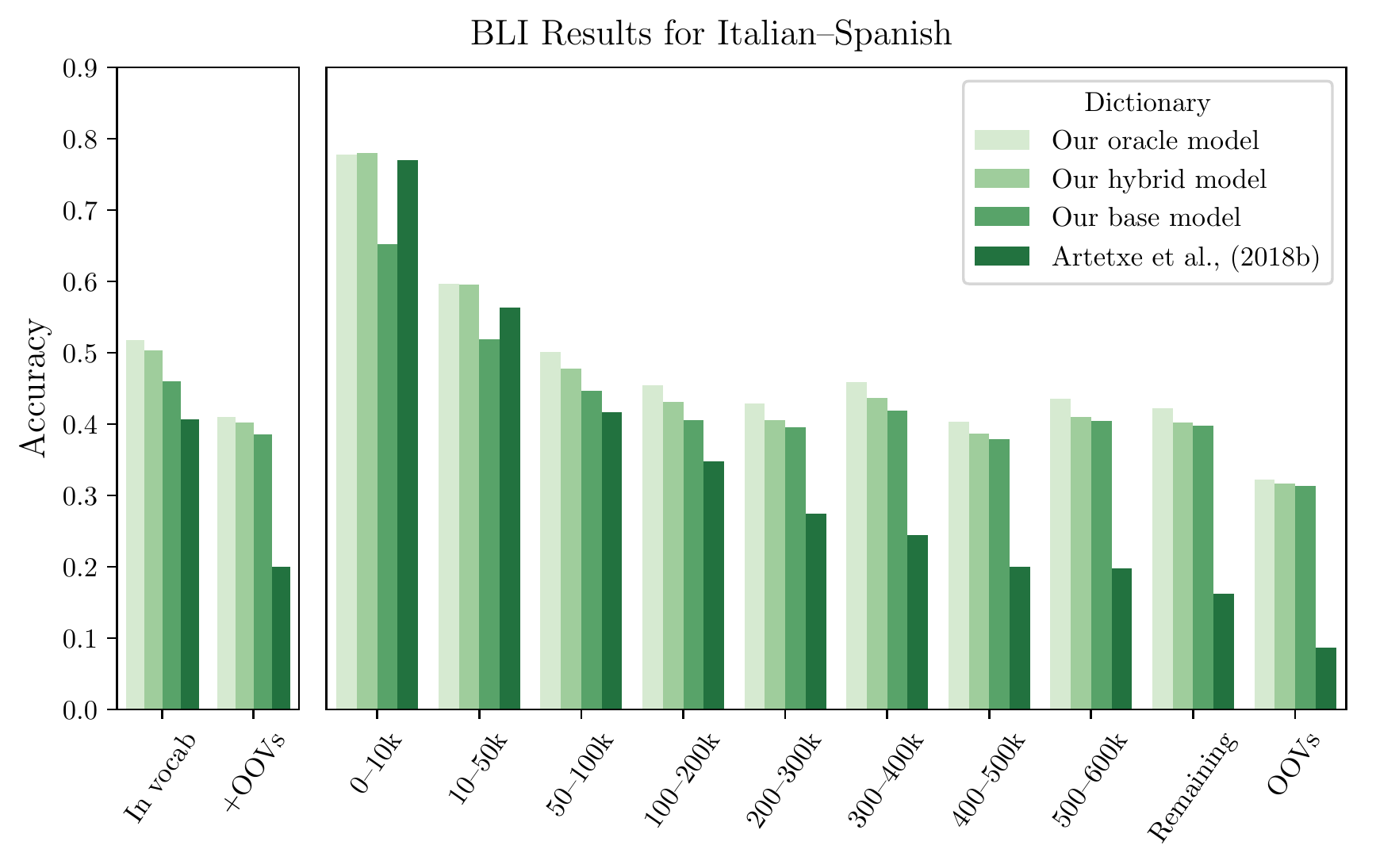}
    \caption{The models' performance on the full dictionary of \newcite{czarnowska2019dont} (left hand side of the plots) and across different word frequency bins, e.g. 0--10k bin is for the top 10k inflected forms.}
    \label{fig:frequency-breakdown}
\end{figure*}

When evaluated on \model{muse} \cite{conneau2018word}, the hybrid model is competitive with the baselines, while our base model performs worse. This is not surprising given that \model{muse} contains only the most frequent inflected forms which, as discussed in \cref{sec:hybrid} and as we demonstrate later in this section, are the weakness of our base model. 
The hybrid model can also suffer from misanalyzing frequent irregular forms of even more frequent lemmata, such as \word{puede} or \word{pudo}---both forms of Spanish \word{poder} (`to be able to').

\paragraph{Frequency Breakdown.}
In \cref{fig:frequency-breakdown} we present the performance breakdown across 10 frequency bins for our base model, its hybrid version, the \R{oracle} and the model of \newcite{artetxe2018robust} for an example language pair: Italian--Spanish. The plot displays results for the supervised setting, for the dictionary of \newcite{czarnowska2019dont}.\footnote{\model{muse} dictionaries \cite{conneau2018word} only contain most frequent forms. 
Thus, we do not show the performance breakdown for that resource.} 
As per the plot, our method proves to be particularly beneficial for translating forms of medium to low frequency; in contrast to the baseline model, for which the performance continuously drops as the forms become less frequent, the performance of our models drops initially, but then plateaus at around 40\% accuracy (for the in-vocabulary forms). 
\cref{fig:frequency-breakdown} also exposes a weakness of our base model---it leads to substantial performance drop for the two highest frequency bins. Notably, the hybrid model does not suffer from this limitation. Indeed, the fact that the shape of the hybrid's plot closely resembles that for the oracle suggests the heuristic we used successfully identifies irregular forms, which are hard to analyze.\looseness=-1
    
\paragraph{Morphology Breakdown.}
In \cref{tab:morph} we present the translation accuracy on French--Spanish, in the supervised setting, for the model of \newcite{artetxe2018robust} and the hybrid model across a range of different morpho-syntactic categories; the models are only evaluated on source forms belonging to a particular paradigm slot. 
We observe that our approach leads to improvements on all but two verbal paradigm slots, while for the adjectives and nouns the performance is competitive with the baselines. 
For many categories the improvements are very substantial; e.g., for the in-vocabulary 2$^{\text{nd}}$ person plural imperative forms of a verb (\T{V;POS;IMP;2;PL}) and 2nd person plural present subjunctive forms (\T{V;SBJV;PRS;2;PL}) the accuracy improves by 55\%.

\begin{table}
\fontsize{8}{10} \selectfont
\renewcommand{\arraystretch}{1}
\parbox{.48\linewidth}{
\begin{tabular}{  l | r r | r r  }
\toprule 
\multirow{2}{*}{Morphology}  & \multicolumn{2}{ c |  }{\makecell{\cite{artetxe2018robust}}} & \multicolumn{2}{ c  }{Our hybrid model}  \\ 
&  \textsc{vocab} & \textsc{all} & \textsc{vocab} & \textsc{all}  \\
\midrule
 \T{adj;fem;pl}      & \W{62.54}   & \W{61.38}    &				58.42 & 57.50    \\	
 \T{adj;fem;sg}      & \W{59.52}   & \W{58.90}    &				57.61 & 57.02    \\	
 \T{adj;masc;pl}       & \W{60.99}   & \W{58.91}    &				60.46 & 58.23    \\	
 \T{adj;masc;sg}       & \W{63.43}   & \W{61.72}    &				61.01 & 59.53    \\	
 \T{n;pl}               & \W{55.21}  & \W{51.81}   &				54.62 & 51.44   \\	
 \T{n;sg}                & \W{57.85}  & \W{56.30}   &				55.26 & 53.79   \\	
\T{v;nfin}            & 48.51   & 47.05    & \W{51.03} & \W{49.45} \\	
\T{v;pos;imp;1;pl}     & 38.13   & 18.02    & \W{64.03} & \W{43.96} \\	
\T{v;pos;imp;2;pl}     & 1.54    & 2.42     & \W{56.92} & \W{45.71} \\	
 \T{v;prs;1;sg}         & 43.14   & 39.78    & \W{43.89} & \W{41.54} \\ 
 \T{v;prs;1;sg}         & 43.14   & 39.78    & \W{43.89} & \W{41.54} \\	
 \T{v;prs;2;pl}         & 0.00     & 2.42     & \W{59.34} & \W{45.71} \\	
 \T{v;prs;2;sg}    2      & 37.05   & 25.71    & \W{48.66} & \W{40.66} \\	
\bottomrule
\end{tabular}
}
\hspace{0.3cm}
\parbox{.48\linewidth}{
\begin{tabular}{  l | r r | r r  }
\toprule 
\multirow{2}{*}{Morphology}  & \multicolumn{2}{ c |  }{\makecell{\cite{artetxe2018robust}}} & \multicolumn{2}{ c  }{Our hybrid model}  \\ 
&  \textsc{vocab} & \textsc{all} & \textsc{vocab} & \textsc{all}  \\
\midrule
 \T{v;prs;3;pl}        & 47.55   & 42.11    & \W{47.80} & \W{43.64} \\
\T{v;prs;3;sg}          & 45.75   & 43.76    & \W{46.21} & \W{44.20} \\
 \T{v;pst;1;sg;ipfv}     & 2.30     & 11.43    & \W{58.62} & \W{46.15} \\	
 \T{v;pst;1;sg;pfv}     & 7.02     & 1.76     & \W{38.60} & \W{43.30} \\	
 \T{v;pst;2;pl;ipfv}    & 0.00      & 0.88     & \W{42.86} & \W{44.84} \\	
 \T{v;pst;2;pl;pfv}    & 0.00      & 0.00     & \W{100.00} & \W{46.81} \\	
 \T{v;pst;3;pl;ipfv}     & \W{60.89}   & 47.59    & {59.41} & \W{48.03} \\	
 \T{v;pst;3;pl;pfv}      & 48.80   & 37.72    & \W{58.80} & \W{47.59} \\	
 \T{v;pst;3;sg;ipfv}     & \W{57.78}   & \W{48.80}    &				55.39 & 47.70    \\	
 \T{v;pst;3;sg;pfv}     & 38.27   & 32.39    & \W{51.96} & \W{46.17} \\	
 \T{v;sbjv;prs;1;pl}     & 20.51    & 4.18     & \W{66.67} & \W{45.93} \\	
\T{v;sbjv;prs;1;sg}   & 43.52   & 39.12    & \W{47.93} & \W{44.18} \\	
\T{v;sbjv;prs;2;pl}     & 0.00     & 0.88     & \W{55.88} & \W{44.62} \\	
\bottomrule
\end{tabular}
}
\caption{Performance on translating source forms belonging to different morpho-syntactic categories for French--Spanish.} 
\label{tab:morph}
 \end{table}

 \section{A Critique of Modern BLI}
 The model we develop in \cref{sec:joint-model} stems from a desire to better integrate inflectional morphology into 
 current state-of-the-art models for BLI. 
 On one hand, the empirical findings we discuss in \cref{sec:results} indicate this attempt was a success, but, on the other, our more nuanced conclusion is that the task of BLI, as currently researched in NLP, is ill-defined {with respect to inflectional morphology}. Indeed, the authors suggest that BLI needs redirection going forward. 
 
{The recent trend in BLI research is to remain data-driven and to avoid specialist linguistic annotation.}
Current projection-based approaches to BLI depend heavily on the assumption that the lexicons of different languages are approximately isomorphic \cite{mikolov2013exploiting,miceli-barone-2016-towards}. 
However, given the immense variation in morphological systems of worlds' languages, this assumption is \textit{prima fascie} false. 
Consider the simple contradiction of Spanish and English, where the first exhibits much more morphological inflection than the latter; there can be no one-to-one alignment between the words in those two lexicons. 
The failure of the isomorphism assumption has been discussed and addressed in many recent works on cross-lingual word embeddings \cite{Søgaard2018,Nakashole2018CharacterizingDF,ormazabal2019analyzing,vulic2019we,Patra2019BilingualLI}. 
However, none of those studies directly target inflectional morphology. In this work we highlight that inflectional morphology complicates BLI and NLP researchers should strive to develop a cleaner way to integrate it into their models. 
We contend the models we present make progress in this direction but there is still a long way to go. 
We now make three concrete suggestions for BLI going forward. 
The first two involve engaging with morphology more seriously and are extensions to the ideas in this paper. 
The third focuses on backing away from morphology.

\paragraph{More Fine-Grained Lexicons.}
Our first proposal is to create more elaborate morphological dictionaries, in the style of those by \newcite{czarnowska2019dont}. 
The primary draw-back of this suggestion is that such resources are tedious to create. \newcite{czarnowska2019dont} focus on genetically related languages, for which inflectional morphological systems are easily compatible. 
However, this is often not the case.  
Every morphological system carves up the semantic space in its own way, e.g. there is no good German equivalent of Polish verbal aspect. 
Thus, such elaborate resources should be carefully crafted and specify lemmata, inflected forms and tags for both the source and the target language. It follows, modulo polysemy and other lexical ambiguity, that the task would be well defined. 

\paragraph{Contextual Word-level Translation.}
Another suggestion for future work is to contextualize word-level translation. Indeed, translation of syntactic features without context is somewhat unusual---e.g. in some contexts a feminine adjective in Spanish might not be translated as feminine in Italian, because a feminine noun in Spanish might be masculine in Italian.
Arguably, much of the morphological ambiguity (and lexical ambiguity) present in modern BLI can be resolved by a word's source-side context.
However, identifying the amount and type of context sufficient to disambiguate possible translations is non-trivial. 
An additional point is that, in this scenario, BLI starts to approach full machine translation. 

\paragraph{Lexeme-Level Translation.}
The final suggestion is for the task of BLI to ignore morphology. 
This would mean filtering the training and test lexicons to ensure that only lemmata exist. 
Such filtering can be performed at the type level with a list of valid lemmata. This third suggestion would entail a return of BLI to the spirit of the task: the induction of a bilingual dictionary. However, creating a lemma-only resource
requires additional language-specific knowledge.

\section{Conclusion} 
We propose a novel model for bilingual lexicon induction which jointly models lexeme translation and inflectional morphology in a structured way. 
Our model improves handling of less frequent, morphologically complex forms, especially if they belong to large inflectional paradigms. 
In our experiments on morphologically enriched dictionaries, we observe substantial performance improvements over the state of the art, particularly prominent for rare verbal paradigm slots. 

\section{Acknowledgments}
{We thank the anonymous reviewers for their thoughtful comments. We would also like to thank Chris Dyer for his helpful suggestions. We acknowledge that Paula is supported by the Cambridge Trust Vice-Chancellor's and Selwyn College scholarship.}

\bibliographystyle{coling}
\bibliography{coling2020}

\newpage
\begin{appendices}

\section{Hyperparameter settings} \label{sec:hyper-params}
\begin{table*}[h]
\centering
\footnotesize
\renewcommand{\arraystretch}{1.1}
\begin{tabular}{ l | c | c | c | c | c | c | c | c  }
\toprule
\multirow{2}{*}{Model}  & \multirowcell{2}{encoder\\layers} & \multirowcell{2}{decoder\\layers} & \multirowcell{2}{encoder\\dim.} & \multirowcell{2}{decoder\\dim.} &  \multirowcell{2}{embedding\\dim.} &  \multirowcell{2}{init. learning\\ rate} & \multirowcell{2}{batch\\ size} & \multirow{2}{*}{dropout} \\
 & & & & & & &   \\
\midrule
Inflector & 2 & 1 & 400 & 400 & 200 & 0.001 &  20 & 0.4  \\
\hline
Analyzer & 2  & 1 & 100 & 100 &  50 & 0.001 & 20 & 0.4  \\
\bottomrule
\end{tabular}
\end{table*}

\begin{table*}[h]
\footnotesize
\centering
\begin{tabular}{  l |  l | l | l |  l | l |  l | l | l | l | l    }
\toprule 
& {fra--ita} &{fra--spa} &{ita--fra} & {ita--spa} & {spa--fra} &{spa--ita} & {eng--ita} & {eng--spa} & {spa--eng} & {ita--eng} \\ 
\hline
batch size & 24 &  24 & 24 & 24 & 24 & 24 & 24 & 24 & 24 & 24 \\
$\alpha$ & 15 & 5 & 15 & 10 & 15 & 10 & 10 & 10 & 10 & 10\\
init. learn. rate & 0.05 & 0.025  & 0.05 & 0.05 & 0.05 & 0.05 &  0.05 & 0.05 & 0.05 & 0.05 \\
\bottomrule
\end{tabular}
\caption{The optimization and hyperparameter settings for the analysis and inflection modules across all language pairs (above) and for the translator module (below). $\alpha$ is the orthogonal regularization weight. {During training we divide the values of $\alpha$ by the batch size to match the weight of the training loss (the losses are averaged across observations for each mini batch).} }

\label{tab:hyperparam}
\end{table*}

\section{Transducer Performance}

\begin{table*}[h]
\footnotesize
\centering
\renewcommand{\arraystretch}{1.3}
\hskip-0.5cm\begin{tabular}{ l |  l | l | l | l | l | l   }
\toprule 
 Task &
 \makecell{\lang{fra}--\lang{ita}} & \makecell{\lang{fra}--\lang{spa}} & \makecell{\lang{ita}--\lang{fra}} & \makecell{\lang{ita}--\lang{spa}} & \makecell{\lang{spa}--\lang{fra}} & \makecell{\lang{spa}--\lang{ita}} \\ 
\hline
 Inflection & 99.1 & 97.4 & 94.2 & 97.9 & 93.6 & 99.1\\ 
\hline
 Analysis & 89.5 & 89.7 & 91.1 & 91.0 & 91.1 & 91.5 \\
\bottomrule 
\end{tabular}
\caption{Transducers' accuracy on the development split of the dictionaries of \newcite{czarnowska2019dont}, when trained on UniMorph entries (excluding those present in the test or development splits of the dictionaries).}
\label{tab:transducers-performance} 
\end{table*}

\noindent \cref{tab:transducers-performance} displays the accuracy of the analysis and inflection components on the development set of the dictionaries of \newcite{czarnowska2019dont}. 
For most language pairs the performance of the inflectors is in the high 90s, while the performance of the analyzers averages at 90.7\%. 
Indeed, the word-type analysis is inherently more difficult than inflection, as it is less deterministic: in most cases there exist more than one correct lemma--tag output. Note that this task is different from the word-token analysis, where the \textit{context} of the analyzed inflection is known. In addition, in contrast to inflection where the model is given the morpho-syntactic tag, the analyzer has no information about the type of word it is handling. {In particular, in an additional line of experimentation, we found that adding a POS to the analyzed source form notably improves the performance. Similarly, training and evaluating on forms associated with a single POS also leads to better accuracy.} 
\end{appendices}
\end{document}